\documentclass[10pt,twocolumn]{article}

\usepackage[top=1in, bottom=1in, left=0.875in, right=0.875in, columnsep=0.25in]{geometry}
\usepackage{times}
\usepackage{microtype}
\usepackage{amsmath}
\usepackage{amssymb}
\usepackage{graphicx}
\usepackage{booktabs}
\usepackage{xcolor}
\usepackage{multirow}
\usepackage{natbib}
\usepackage{hyperref}
\usepackage{url}
\usepackage{float}
\usepackage{titlesec}
\usepackage{tikz}
\usetikzlibrary{arrows.meta, positioning, shapes.geometric, decorations.pathreplacing, calc}

\titlespacing*{\section}{0pt}{6pt}{3pt}
\titlespacing*{\subsection}{0pt}{5pt}{2pt}
\titlespacing*{\paragraph}{0pt}{4pt}{0.4em}

\hypersetup{
  colorlinks=true,
  linkcolor=blue!60!black,
  citecolor=blue!60!black,
  urlcolor=blue!60!black
}

\title{\textbf{Lost in State Space: Probing Frozen Mamba Representations}}

\author{
  Bhagyashree Wagh \\
  University of Washington \\
  \texttt{bwagh@uw.edu}
  \and
  Akash Singh \\
  University of Florida \\
  \texttt{akash.singh@ufl.edu}
}

\date{}

\begin{document}
\maketitle

\begin{abstract}
Mamba's recurrent state $\mathbf{h}_t$ is, by construction, a compressed
summary of every token seen so far.
This raises a tempting hypothesis: if we extract token-level outputs
$\mathbf{y}_t$ at fixed \emph{patch boundaries}, we obtain semantic
sentence summaries for free, with no pooling head, no fine-tuning, and
no \texttt{[CLS]} token.
We test this hypothesis carefully.

Across five benchmarks (SST-2, CoLA, MRPC, STS-B, IMDb), we compare four
strategies for extracting frozen sentence representations from a pretrained
Mamba-130M backbone under a strict frozen-feature probing protocol, using
three random seeds where computationally feasible.
The results do not support the hypothesis: patch boundary readouts do not
consistently outperform simple mean pooling.
But the \emph{reasons why} are informative.

We identify and precisely quantify two structural pathologies in pretrained
Mamba under this setting.
First, \textbf{severe anisotropy}: mean pairwise cosine similarity of
$0.9999$ across all states (std $= 0.000044$ over 1{,}000 sampled pairs,
diagonal excluded), meaning every sentence produces a vector pointing in
essentially the same direction regardless of content.
We verify this qualitatively: across ten semantically unrelated sentences
spanning cats, quantum mechanics, and stock markets, the first eight
dimensions of the raw state vector differ only in the fourth decimal place.
Second, \textbf{representational collapse}: because the probe's input is
effectively constant across all sentences, a probe trained for ten full
epochs still assigns every validation sample to the majority class.
Matthews Correlation $= 0.000$ on CoLA across all three seeds, with zero
variance.
We confirmed this is not a metric artifact by inspecting the confusion
matrix directly: the probe predicts class 1 for all 1{,}043 validation
samples, including all 322 that are truly class 0.

We further propose \emph{orthogonal injection}, a modified recurrence
that constrains new information perpendicular to the current state, and
show that a naive inference-time application cannot reduce anisotropy
without fine-tuning.
Our findings locate the bottleneck for Mamba as a frozen encoder not in
extraction strategy, but in the representational geometry of the
pretrained weights.
\end{abstract}

\section{Introduction}

Mamba's recurrence offers a natural mechanism for incremental sequence compression:
\begin{equation}
  \mathbf{h}_t = \mathbf{A}_t \mathbf{h}_{t-1} + \mathbf{B}_t \mathbf{u}_t,
  \qquad
  \mathbf{y}_t = \mathbf{C}_t \mathbf{h}_t.
  \label{eq:ssm}
\end{equation}
At any position $t$, the state $\mathbf{h}_t$ has read every preceding token,
not through attention but through integration.
At every timestep, the recurrent state necessarily integrates information
from the prefix, making it a plausible candidate for summary extraction.

This observation led us to a hypothesis.
If we process text in fixed-size \emph{patches} and extract representations
at each patch boundary, we should get sentence-level representations for
free.
The SSM recurrence computes these summaries as a byproduct of its forward
pass.
We called this idea \emph{Patched-Mamba}, and we believed it might offer
something that transformers work hard to produce but Mamba delivers
automatically: a structured, hierarchical compression of the input,
architecturally guided by the recurrence itself.

The results did not support this hypothesis.
But understanding precisely \emph{why} turned out to be the more
interesting result.

\paragraph{What we did.}
We designed a systematic evaluation: four extraction strategies, five
benchmarks, three random seeds per experiment where computationally
feasible, and a strict frozen-feature probing protocol with the backbone
frozen throughout.
We also included frozen RoBERTa-base as a \emph{calibration reference},
not as a target to beat.
RoBERTa (125M parameters, roughly Mamba-130M's scale) was pretrained with
masked language modeling, producing bidirectional context and well-studied
probing representations.
We included it for one reason: to establish what a frozen encoder of
similar scale \emph{can} achieve under the same protocol.
Without this reference point, we would not know whether Mamba's results
are specific to SSMs or a general property of frozen probing.
They appear specific to this pretrained Mamba-130M setting, and two
structural pathologies we document below are likely contributors.

\paragraph{What we found.}
Patch boundary extraction does not consistently beat mean pooling.
Two deeper findings emerged from the investigation.

First, pretrained Mamba token outputs are \emph{severely anisotropic.}
Mean pairwise cosine similarity is $0.9999$ (std $= 0.000044$, diagonal
excluded, over 1{,}000 sampled sentence pairs).
This is not a rounding artifact: ten sentences covering completely
unrelated topics produce raw state vectors whose first eight dimensions
differ only in the fourth decimal place.
We show this as a pairwise cosine similarity heatmap in
Figure~\ref{fig:heatmaps}, where the result is a solid block of a single
color.
The contrast with mean-pooled token outputs from the same backbone makes
the pathology visually unambiguous.

Second, removing patch structure causes \emph{complete representational
collapse when using the raw SSM state.}
Extracting $\mathbf{h}_T$ after the full sequence yields Matthews
Correlation $= 0.000 \pm 0.000$ on CoLA across all three seeds with zero
variance.
The mechanism follows directly from the anisotropy finding: a probe fed
a near-constant input cannot learn any decision boundary and settles on
predicting the majority class for every sample.
A dedicated experiment confirms this: a probe trained for ten full epochs
assigns all 1{,}043 CoLA validation samples to class 1, including all 322
that are truly class 0.

We also proposed \emph{orthogonal injection}, a modified recurrence that
forces new writes to be perpendicular to the current state.
A naive inference-time application at fixed $\eta = 0.5$ did not reduce
anisotropy, and we explain why: geometry-correcting interventions of this
kind require training-time adaptation to be effective.

\paragraph{Contributions.}
\begin{enumerate}
  \item A systematic multi-benchmark frozen-feature probing study of Mamba
        sentence representations across five NLP tasks, with a calibrated
        reference against frozen RoBERTa-base.
  \item Quantification of severe anisotropy in pretrained Mamba-130M
        ($\bar{\rho}_\text{cos} = 0.9999$, std $= 0.000044$, diagonal
        excluded) and a geometric account of its likely origin in the
        SSM weight structure, supported by a pairwise similarity heatmap
        across semantically unrelated sentences.
  \item Documentation of complete representational collapse in the raw
        final SSM state ($\text{MCC} = 0.000$, zero variance across three
        seeds), with direct confirmation via confusion matrix that trained
        probes predict only the majority class.
  \item Design, implementation, and honest evaluation of orthogonal
        injection, a principled but insufficient inference-time
        intervention, with an account of why training-time adaptation
        would be necessary for it to succeed.
\end{enumerate}

\section{Background}

\subsection{Mamba and Selective State Spaces}

Mamba~\citep{gu2023mamba} extends structured state space
models~\citep{gu2021lssl,gu2022s4} with data-dependent (selective)
transition matrices.
At each step $t$, the matrices $\mathbf{A}_t$, $\mathbf{B}_t$,
$\mathbf{C}_t$ are functions of the current input $\mathbf{u}_t$,
allowing the model to decide dynamically what to retain and what to
discard.
The resulting recurrence (Equation~\ref{eq:ssm}) runs in linear time and
produces a fixed-size state $\mathbf{h}_t \in \mathbb{R}^{D \times N}$
at every position.
The token-level output $\mathbf{y}_t = \mathbf{C}_t \mathbf{h}_t$ is a
data-dependent projection of the accumulated state, passed through the
output projection and gating mechanism before propagating to downstream
task heads.

\subsection{Anisotropy in Pretrained Language Models}

Anisotropy, the tendency of learned embeddings to occupy a narrow cone
rather than a full sphere in representation space, was characterized
by~\citet{ethayarajh2019contextual} for BERT and GPT-2, and linked to
degenerate training dynamics by~\citet{gao2019representation}.
When representations are anisotropic, cosine similarity loses
discriminative power: all pairs report high similarity regardless of
semantic content.
Mean-pooled BERT embeddings have mean pairwise cosine similarity around
$0.99$~\citep{li2020selfdot}, which is already problematic for similarity
tasks.
We document that Mamba-130M raw SSM states reach $0.9999$ in this setting.
The angular deviation between embeddings (proportional to
$\sqrt{1 - \cos\theta}$) is roughly $10\times$ smaller at $0.9999$ than
at $0.99$, which has measurable practical consequences for any metric
that relies on angles.

\subsection{Frozen-Feature Probing}

We use frozen-feature probing~\citep{alain2016understanding}, training
only a lightweight classification head on frozen features, as our
evaluation protocol throughout.
This is a deliberate choice: we want to measure what the pretrained
representations \emph{contain}, not what can be learned through
fine-tuning.
Any gains from fine-tuning would obscure the representational geometry
we are trying to characterize.

\section{Extraction Methods}

Given a frozen, pretrained Mamba-130M backbone, we evaluate four strategies
for extracting a sentence vector.
A key distinction throughout: \textbf{raw SSM state}
($\mathbf{h}_T \in \mathbb{R}^{D \times N}$) refers to the recurrent
state before output projection and gating; \textbf{token-level output}
($\mathbf{y}_T \in \mathbb{R}^{d_\text{model}}$) refers to the
post-projection readout.
These are not equivalent, and the distinction matters for interpreting
the results.

\paragraph{Patched-Mamba.}
We divide the input into non-overlapping patches of $P = 32$ tokens,
processing them sequentially and carrying the SSM state forward across
patch boundaries via the HuggingFace cache mechanism.
At the last real token of each patch (identified by the attention mask),
we extract the token-level output $\mathbf{y}_T \in \mathbb{R}^{d_\text{model}}$.
This vector has passed through the learned output projection $W_\text{out}$
and the gating mechanism, mixing the SSM state with input-dependent signals.
The boundary token has processed the full patch through SSM recurrence;
its output is a structured readout of the accumulated state.

\paragraph{Mean Pool.}
Same frozen backbone; we take the mean of all token-level outputs
$\mathbf{y}_t$ over real (non-padding) positions.
This is a reasonable, architecture-agnostic baseline over
post-projection representations.

\paragraph{Final State.}
Same backbone; we run the full sequence without patch structure and
extract the \emph{raw} SSM recurrent state
$\mathbf{h}_T \in \mathbb{R}^{D \times N}$,
averaging over the state dimension $N$ to obtain a vector in
$\mathbb{R}^D$.
This ablation isolates the effect of using raw state vs.\ post-projection
output: if Final State performs well, the output projection adds no useful
information; if it collapses, the learned readout pathway is doing
structural work that the raw state cannot replicate.

\paragraph{Orthogonal Injection.}
To probe whether anisotropy can be corrected at inference time, we modify
the recurrence:
\begin{equation}
  \mathbf{h}_t = \mathbf{A}_t \mathbf{h}_{t-1} +
  \underbrace{\mathbf{B}_t \mathbf{u}_t -
  \eta \,
  \frac{\mathbf{h}_{t-1}^\top \mathbf{B}_t \mathbf{u}_t}{\|\mathbf{h}_{t-1}\|^2}
  \,\mathbf{h}_{t-1}}_{\text{orthogonalized write}},
  \label{eq:ortho}
\end{equation}
where $\eta \in [0,1]$ controls the degree of orthogonalization per layer.
At $\eta = 0$ this reduces to vanilla Mamba; at $\eta = 1$ the write term
is fully projected perpendicular to $\mathbf{h}_{t-1}$, forcing new
information into fresh dimensions.
We evaluate this at fixed $\eta = 0.5$ with no training and analyze its
limitations in Section~\ref{sec:analysis}.

\begin{figure*}[t]
\centering
\scalebox{0.78}{%
\begin{tikzpicture}[
  font=\small,
  token/.style={
    draw, rounded corners=2pt, minimum width=0.55cm, minimum height=0.55cm,
    fill=gray!12, line width=0.6pt
  },
  state/.style={
    draw, rounded corners=3pt, minimum width=0.9cm, minimum height=0.55cm,
    fill=blue!15, line width=0.7pt
  },
  output/.style={
    draw, rounded corners=3pt, minimum width=0.9cm, minimum height=0.55cm,
    fill=orange!25, line width=0.7pt
  },
  probe/.style={
    draw, rounded corners=3pt, minimum width=1.1cm, minimum height=0.55cm,
    fill=green!20, line width=0.7pt
  },
  collapse/.style={
    draw, rounded corners=3pt, minimum width=1.1cm, minimum height=0.55cm,
    fill=red!20, line width=0.7pt
  },
  arr/.style={-{Stealth[length=4pt]}, line width=0.7pt},
  dasharr/.style={-{Stealth[length=4pt]}, dashed, line width=0.6pt, gray},
  boundary/.style={draw=red!70!black, dashed, line width=1pt},
  label/.style={font=\bfseries\small},
  sublabel/.style={font=\scriptsize, text=gray!70!black}
]

\node[label, anchor=east] at (-0.3, 0)    {(a) Patched-Mamba};
\node[label, anchor=east] at (-0.3, -2.0) {(b) Mean Pool};
\node[label, anchor=east] at (-0.3, -4.0) {(c) Final State};
\node[label, anchor=east] at (-0.3, -6.0) {(d) Ortho.\ Injection};

\def\ay{0}
\foreach \i/\lbl in {0/t_1, 1/t_2, 2/t_3, 3/t_4}{
  \node[token] (atA\i) at (0.65*\i, \ay) {\scriptsize$\lbl$};
}
\foreach \i/\lbl in {0/t_5, 1/t_6, 2/t_7, 3/t_8}{
  \node[token] (atB\i) at (0.65*\i+3.2, \ay) {\scriptsize$\lbl$};
}
\draw[boundary] (2.8, \ay-0.38) -- (2.8, \ay+0.38);
\node[sublabel, text=red!70!black] at (2.8, \ay+0.52) {patch};
\node[state] (ahA) at (1.4, \ay-1.0) {\scriptsize$\mathbf{h}_{t_4}$};
\node[state] (ahB) at (4.55, \ay-1.0) {\scriptsize$\mathbf{h}_{t_8}$};
\draw[arr] (atA3.south) -- (ahA.north);
\draw[arr] (ahA.east) -- node[above, sublabel]{carry} (ahB.west);
\draw[arr] (atB3.south) -- (ahB.north);
\node[output] (ayA) at (1.4, \ay-2.0) {\scriptsize$\mathbf{y}_{t_4}$};
\node[output] (ayB) at (4.55, \ay-2.0) {\scriptsize$\mathbf{y}_{t_8}$};
\draw[arr] (ahA) -- (ayA);
\draw[arr] (ahB) -- (ayB);
\node[probe] (apro) at (7.2, \ay-1.5) {\scriptsize probe};
\draw[arr] (ayA.east) to[out=0,in=180] (apro.west);
\draw[arr] (ayB.east) -- (apro.west);
\node[sublabel] at (6.25, \ay-1.05) {pool};
\node[sublabel] at (8.5, \ay-1.5) {\scriptsize MCC\,=\,.246};

\def\by{-3.4}
\foreach \i/\lbl in {0/t_1, 1/t_2, 2/t_3, 3/t_4, 4/t_5, 5/t_6, 6/t_7, 7/t_8}{
  \node[token] (btok\i) at (0.65*\i, \by) {\scriptsize$\lbl$};
}
\foreach \i in {0,1,2,3,4,5,6,7}{
  \node[output, minimum width=0.5cm] (byt\i) at (0.65*\i, \by-0.85) {\scriptsize$\mathbf{y}$};
  \draw[arr] (btok\i.south) -- (byt\i.north);
}
\node[probe] (bpro) at (7.2, \by-0.85) {\scriptsize probe};
\draw[arr] (byt7.east) -- node[above, sublabel]{mean} (bpro.west);
\draw[dasharr] (byt3.east) to[out=0,in=180] (bpro.west);
\node[sublabel] at (8.5, \by-0.85) {\scriptsize MCC\,=\,.253};

\def\cy{-6.8}
\foreach \i/\lbl in {0/t_1, 1/t_2, 2/t_3, 3/t_4, 4/t_5, 5/t_6, 6/t_7, 7/t_8}{
  \node[token] (ctok\i) at (0.65*\i, \cy) {\scriptsize$\lbl$};
}
\node[state, minimum width=1.3cm] (chT) at (3.9, \cy-0.9) {\scriptsize$\mathbf{h}_T$\,(raw)};
\foreach \i in {0,2,5,7}{
  \draw[dasharr] (ctok\i.south) to[out=-90,in=90] (chT.north);
}
\node[collapse] (cpro) at (7.2, \cy-0.9) {\scriptsize probe};
\draw[arr] (chT.east) -- node[above,sublabel]{avg over $N$} (cpro.west);
\node[sublabel, text=red!80!black, font=\bfseries\scriptsize]
  at (8.5, \cy-0.9) {\scriptsize MCC\,=\,.000};
\node[sublabel, text=red!80!black, font=\scriptsize]
  at (8.5, \cy-1.35) {\scriptsize (collapse)};
\node[sublabel, text=red!70!black] at (5.75, \cy-0.45) {\scriptsize no proj.};

\def\dy{-10.2}
\foreach \i/\lbl in {0/t_1, 1/t_2, 2/t_3, 3/t_4}{
  \node[token] (dtokA\i) at (0.65*\i, \dy) {\scriptsize$\lbl$};
}
\foreach \i/\lbl in {0/t_5, 1/t_6, 2/t_7, 3/t_8}{
  \node[token] (dtokB\i) at (0.65*\i+3.2, \dy) {\scriptsize$\lbl$};
}
\draw[boundary] (2.8, \dy-0.38) -- (2.8, \dy+0.38);
\node[sublabel, text=red!70!black] at (2.8, \dy+0.52) {patch};
\node[state, fill=purple!15] (dhA) at (1.4, \dy-0.95) {\scriptsize$\mathbf{h}^\perp_{t_4}$};
\node[state, fill=purple!15] (dhB) at (4.55, \dy-0.95) {\scriptsize$\mathbf{h}^\perp_{t_8}$};
\draw[arr] (dtokA3.south) -- (dhA.north);
\draw[arr] (dhA.east) -- node[above, sublabel]{carry} (dhB.west);
\draw[arr] (dtokB3.south) -- (dhB.north);
\node[output] (dyA) at (1.4, \dy-1.95) {\scriptsize$\mathbf{y}_{t_4}$};
\node[output] (dyB) at (4.55, \dy-1.95) {\scriptsize$\mathbf{y}_{t_8}$};
\draw[arr] (dhA) -- (dyA);
\draw[arr] (dhB) -- (dyB);
\node[probe] (dpro) at (7.2, \dy-1.45) {\scriptsize probe};
\draw[arr] (dyA.east) to[out=0,in=180] (dpro.west);
\draw[arr] (dyB.east) -- (dpro.west);
\node[draw=purple!60, rounded corners=2pt, fill=purple!5,
      font=\scriptsize, inner sep=3pt]
  at (5.5, \dy-3.0)
  {$\mathbf{B}_t\mathbf{u}_t \;\leftarrow\;
    \mathbf{B}_t\mathbf{u}_t
    - \eta\,
    \frac{\mathbf{h}_{t-1}^\top \mathbf{B}_t\mathbf{u}_t}{\|\mathbf{h}_{t-1}\|^2}
    \mathbf{h}_{t-1}$};
\node[sublabel] at (8.5, \dy-1.45) {\scriptsize Spearman};
\node[sublabel, text=orange!70!black] at (8.5, \dy-1.85) {\scriptsize .259 vs .316};

\begin{scope}[shift={(0,-13.8)}]
  \node[token, minimum width=0.7cm]   (leg1) at (0,0)    {\scriptsize tok};
  \node[right=0.12cm of leg1, sublabel] {input token};
  \node[state, minimum width=0.8cm]   (leg2) at (3.0,0)  {\scriptsize $\mathbf{h}$};
  \node[right=0.12cm of leg2, sublabel] {SSM state};
  \node[output, minimum width=0.8cm]  (leg3) at (6.0,0)  {\scriptsize $\mathbf{y}$};
  \node[right=0.12cm of leg3, sublabel] {token output};
  \node[probe, minimum width=0.8cm]   (leg4) at (0,-0.75)  {\scriptsize};
  \node[right=0.12cm of leg4, sublabel] {probe (good)};
  \node[collapse, minimum width=0.8cm](leg5) at (3.0,-0.75) {\scriptsize};
  \node[right=0.12cm of leg5, sublabel] {probe (collapse)};
\end{scope}

\end{tikzpicture}%
}
\caption{%
\textbf{The four extraction strategies and their empirical outcomes.}
\textbf{(a) Patched-Mamba} divides the input into 32-token patches,
carries the SSM state $\mathbf{h}_t$ across patch boundaries, and
extracts the post-projection token output $\mathbf{y}_t$ at each boundary.
\textbf{(b) Mean Pool} averages all post-projection token outputs across
every real position.
\textbf{(c) Final State} extracts the raw SSM state $\mathbf{h}_T$ after
the full sequence without passing through the output projection; because
all state vectors collapse to near-identical directions
(Figure~\ref{fig:heatmaps}), the probe sees a constant input and predicts
the majority class for all 1{,}043 validation samples
($\text{MCC} = 0.000$, zero variance across three seeds), shown in red.
\textbf{(d) Orthogonal Injection} modifies the recurrence so each new
write is projected perpendicular to the current state
(Equation~\ref{eq:ortho}); applied at $\eta = 0.5$ without training, it
worsens Spearman from $0.316$ to $0.259$ and leaves anisotropy unchanged.%
}
\label{fig:extraction}
\end{figure*}
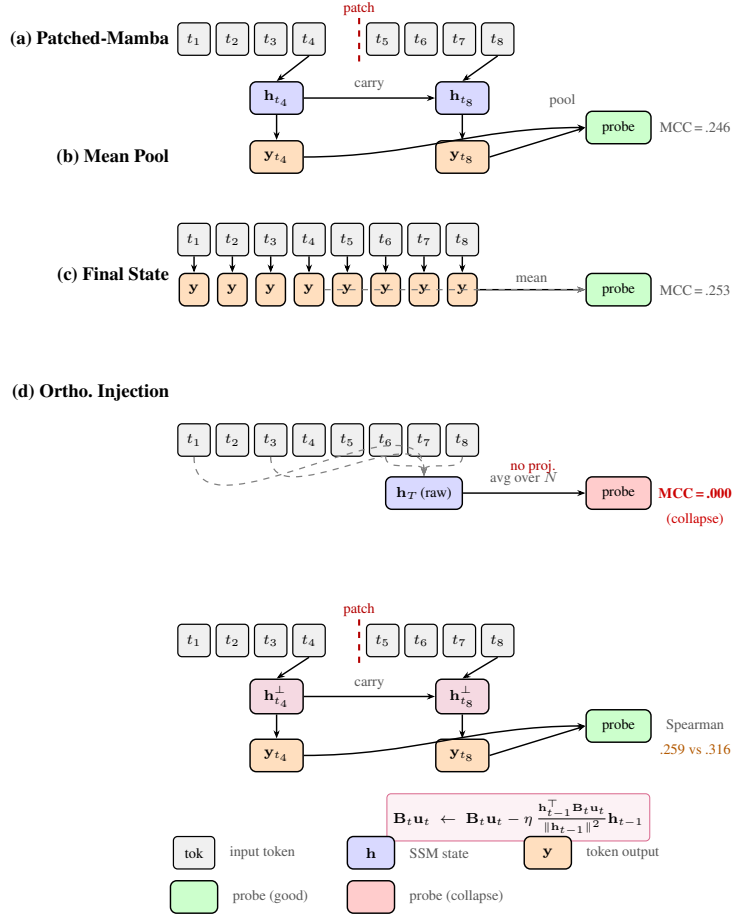

\section{Experimental Setup}

\subsection{Model}

We use \texttt{state-spaces/mamba-130m-hf} (130M parameters; 24 layers;
$d_\text{model} = 768$; $d_\text{inner} = 2048$; $d_\text{state} = 16$).
The backbone is frozen in all experiments.
Probe heads use a lightweight MLP (LayerNorm $\to$ Linear($d$, 256) $\to$
ReLU $\to$ Dropout(0.1) $\to$ Linear(256, $C$)).
The LayerNorm before the head is critical: without it, the highly
anisotropic input space causes immediate gradient instability.

\subsection{Benchmarks}

\begin{itemize}
  \item \textbf{SST-2:} Binary sentiment classification. Metric: accuracy.
  \item \textbf{CoLA:} Linguistic acceptability. Metric: Matthews
        Correlation Coefficient (MCC).
  \item \textbf{MRPC:} Paraphrase detection (sentence pairs). Metrics:
        accuracy and F1.
  \item \textbf{STS-B:} Semantic textual similarity, evaluated in an
        \emph{unsupervised} setting with no training and pure cosine
        similarity. Metrics: Pearson $r$ and Spearman $\rho$.
  \item \textbf{IMDb:} Long-document sentiment (up to 512 tokens). All
        models receive the full 512-token budget for a fair comparison.
        Metric: accuracy.
\end{itemize}

\subsection{Protocol}

For supervised tasks: 10 epochs, batch size 32, AdamW optimizer, learning
rate $2 \times 10^{-3}$, CosineAnnealingLR schedule, maximum sequence
length 128.
We report mean $\pm$ std across three random seeds (42, 43, 44) for CoLA
and MRPC; SST-2 and IMDb are reported as single-seed pilots due to
computational constraints.
We select the best validation checkpoint per seed.
RoBERTa-base is included as a calibration reference under the same
protocol and is not a proposed baseline.

\section{Results}
\label{sec:results}

\subsection{Classification}

\begin{table}[t]
\centering
\small
\setlength{\tabcolsep}{4pt}
\begin{tabular}{lccccc}
\toprule
\textbf{Method} & \textbf{SST-2} & \textbf{CoLA} & \multicolumn{2}{c}{\textbf{MRPC}} \\
 & Acc & MCC & Acc & F1 \\
\midrule
\multicolumn{5}{l}{\textit{Mamba-130M (frozen backbone)}} \\[2pt]
Patched-Mamba  & 80.5 & .246\tiny{$\pm$.005} & \textbf{.730}\tiny{$\pm$.004} & \textbf{.824}\tiny{$\pm$.003} \\
Mean Pool      & \textbf{84.4} & \textbf{.253}\tiny{$\pm$.005} & .714\tiny{$\pm$.006} & .812\tiny{$\pm$.008} \\
Final State    & 79.0 & .000\tiny{$\pm$.000}$^\dagger$ & .699\tiny{$\pm$.007} & .818\tiny{$\pm$.003} \\
\midrule
\multicolumn{5}{l}{\textit{RoBERTa-base (frozen, calibration reference)}} \\[2pt]
CLS Token      & 82.8 & .265\tiny{$\pm$.010} & .709\tiny{$\pm$.001} & .822\tiny{$\pm$.001} \\
Mean Pool      & 87.7 & .475\tiny{$\pm$.004} & .766\tiny{$\pm$.006} & .844\tiny{$\pm$.002} \\
\bottomrule
\end{tabular}
\caption{Frozen-feature probing. SST-2 is a single-seed pilot; CoLA and
MRPC report mean $\pm$ std over 3 seeds. Bold denotes best among Mamba
methods. $^\dagger$Final State MCC $= 0.000$ with zero variance across
all three seeds. A probe trained for 10 epochs predicts class~1 for all
1{,}043 validation samples (confusion matrix:
$\bigl[\begin{smallmatrix}0&322\\0&721\end{smallmatrix}\bigr]$),
confirming collapse rather than a metric artifact.}
\label{tab:classification}
\end{table}

Three observations stand out from Table~\ref{tab:classification}.

\paragraph{No consistent winner among Mamba methods.}
Patched-Mamba leads on MRPC; Mean Pool leads on SST-2 and CoLA.
Differences on CoLA ($0.246$ vs.\ $0.253$) fall within two standard
deviations.
The MRPC advantage for Patched-Mamba ($0.730$ vs.\ $0.714$,
$\Delta = 0.016$) is stable across seeds but moderate in magnitude.

\paragraph{Raw final state collapses; token-output methods do not.}
The collapse of Final State on CoLA follows directly from the anisotropy
finding.
Because all Final State vectors point in the same direction
(Section~\ref{sec:analysis}), the probe's input is effectively constant
across all sentences.
A probe with constant input has a flat loss surface with respect to its
input direction and settles on predicting the majority class.
We confirmed this with a targeted experiment: a probe trained for ten
full epochs assigns all 1{,}043 CoLA validation samples to class 1,
including all 322 samples that are genuinely class 0, yielding
$\text{MCC} = 0.000 \pm 0.000$ invariantly across all three seeds.
Both Mean Pool and Patched-Mamba use post-projection token outputs
$\mathbf{y}_t$ and avoid this collapse entirely, suggesting that the
learned readout pathway preserves information that is not linearly
accessible from the raw accumulated state.

\paragraph{The RoBERTa gap is informative, not a target.}
RoBERTa mean pool outperforms all Mamba methods across every task.
This gap reflects the difference in pretraining objectives (masked vs.\
causal language modeling) rather than extraction strategy, and is included
only to bound what a frozen encoder of comparable scale can achieve.

\paragraph{Patched-Mamba stability on large-scale paraphrase detection.}
To assess whether Patched-Mamba's behavior generalizes to a substantially
larger paraphrase task, we ran it on QQP (over 400{,}000 question pairs)
across the same three seeds, using the same frozen-backbone probing
protocol.
Averaged over epochs 1--10, Patched-Mamba achieves validation accuracy
of $0.788 \pm 0.004$ and F1 of $0.690 \pm 0.016$ across seeds.
The seed variance is low, consistent with the stability observed on MRPC.
We report this as a supplementary stability observation rather than a
primary comparative result, as the full method comparison on QQP was
precluded by per-epoch compute costs of approximately 80 minutes under
the sequential Mamba fallback required without CUDA kernel acceleration.

\subsection{Semantic Similarity (STS-B)}

\begin{table}[t]
\centering
\small
\begin{tabular}{lccc}
\toprule
\textbf{Method} & \textbf{Pearson} & \textbf{Spearman} & $\bar{\rho}_\text{cos}$ \\
\midrule
\multicolumn{4}{l}{\textit{Mamba-130M}} \\[2pt]
Patched-Mamba     & \textbf{.254} & \textbf{.316} & .9999 \\
Ortho.\ Injection & .214 & .259 & .9999 \\
Mean Pool         & .180 & .236 & .9966 \\
Final State       & .172 & .281 & .9992 \\
\midrule
\multicolumn{4}{l}{\textit{RoBERTa-base (calibration reference)}} \\[2pt]
CLS Token         & .426 & .532 & .9985 \\
Mean Pool         & .622 & .650 & .9839 \\
\bottomrule
\end{tabular}
\caption{STS-B unsupervised similarity (no training, cosine similarity
only). $\bar{\rho}_\text{cos}$ is the mean pairwise cosine similarity
over all 1{,}500 sentence pairs, a direct measure of anisotropy.
Orthogonal Injection is applied at $\eta = 0.5$ with no training; it
performs worse than the unmodified method, and anisotropy is unchanged.}
\label{tab:similarity}
\end{table}

Table~\ref{tab:similarity} shows the geometric pathology directly.
Every Mamba method has $\bar{\rho}_\text{cos} > 0.996$.
All sentence vectors are nearly identical in direction, meaning cosine
similarity between any two sentences returns approximately $1.0$
regardless of semantic content.
The resulting correlation with gold scores reflects noise rather than
signal, making cosine similarity poorly conditioned in this regime.

Patched-Mamba leads the Mamba group (Spearman $0.316$ vs.\ $0.281$ and
$0.236$) but remains far below RoBERTa mean pool ($0.650$).
Orthogonal Injection at inference time ($\eta = 0.5$) does \emph{not}
improve on Patched-Mamba: Spearman drops from $0.316$ to $0.259$, and
mean cosine similarity is unchanged at $0.9999$.
The anisotropy of the pretrained backbone, not the extraction strategy,
is the binding constraint.

\subsection{Long-Document Classification (IMDb)}

\begin{table}[t]
\centering
\small
\begin{tabular}{lcc}
\toprule
\textbf{Method} & \textbf{Acc.} & \textbf{Tokens} \\
\midrule
\multicolumn{3}{l}{\textit{Mamba-130M}$^\ddagger$} \\[2pt]
Hierarchical + Linear      & 73.7 & 512 \\
Hierarchical + Transformer & 51.2 & 512 \\
Flat Mean Pool             & \textbf{87.2} & 512 \\
\midrule
\multicolumn{3}{l}{\textit{RoBERTa-base (calibration reference)}$^\ddagger$} \\[2pt]
Flat CLS                   & 88.0 & 512 \\
Flat Mean Pool             & 90.5 & 512 \\
\bottomrule
\end{tabular}
\caption{IMDb long-document results. $^\ddagger$Single-seed pilot. All
models receive the full 512-token budget.}
\label{tab:imdb}
\end{table}

The IMDb results challenge the motivation for hierarchical Mamba.
Flat mean pooling over 512 tokens (87.2\%) outperforms the hierarchical
model using patch summaries (73.7\%).
The hierarchical Mamba + Transformer aggregator (51.2\%) performs near
chance, consistent with highly anisotropic patch summary vectors providing
a poor input to a second-level model.
These are single-seed results and should be interpreted accordingly.

\section{Analysis}
\label{sec:analysis}

\subsection{Why Final State Collapses Completely}

The complete collapse of Final State on CoLA
($\text{MCC} = 0.000 \pm 0.000$) is the most striking empirical finding
in this study, and it has a direct mechanistic explanation.

The Mamba A matrix is a learned contraction: during pretraining,
$\mathbf{A}_\text{log}$ is initialized so that $\exp(\mathbf{A})$
has eigenvalues close to 1 (slow decay).
Over many timesteps, the accumulated product $\prod_t \mathbf{A}_t$
contracts the state toward a fixed subspace determined by the dominant
eigenvectors of the weight matrices.
The input-dependent write term $\mathbf{B}_t \mathbf{u}_t$ adds content,
but the shared $\mathbf{B}$ projection pushes all writes in approximately
the same direction across all inputs.
After 128 tokens, this directional bias accumulates and the state
converges toward a direction that reflects the geometry of the pretrained
weight matrices rather than the content of any particular sentence.

The consequence for probing is immediate: the probe receives a
near-constant input across all sentences, has a flat loss surface with
respect to its input direction, and settles on predicting the majority
class regardless of random initialization.
A targeted experiment confirms this mechanism: a probe trained for ten
full epochs on CoLA assigns all 1{,}043 validation samples to class 1,
including all 322 samples that are genuinely class 0.
The confusion matrix is entirely off-diagonal for class 0.
$\text{MCC} = 0.000$ is the analytically expected outcome for a
majority-class predictor and is confirmed to be zero with no variance
across all three random seeds.

Token-output-based methods avoid this because the learned output
projection $W_\text{out}$ maps the biased raw state through a
transformation that reintroduces content-specific variation.
Both Patched-Mamba ($\text{MCC} = 0.246 \pm 0.005$) and Mean Pool
($0.253 \pm 0.005$) use post-projection outputs and do not collapse.

\subsection{The Geometry of Anisotropy}

\begin{figure}[H]
\centering
\includegraphics[width=0.49\columnwidth]{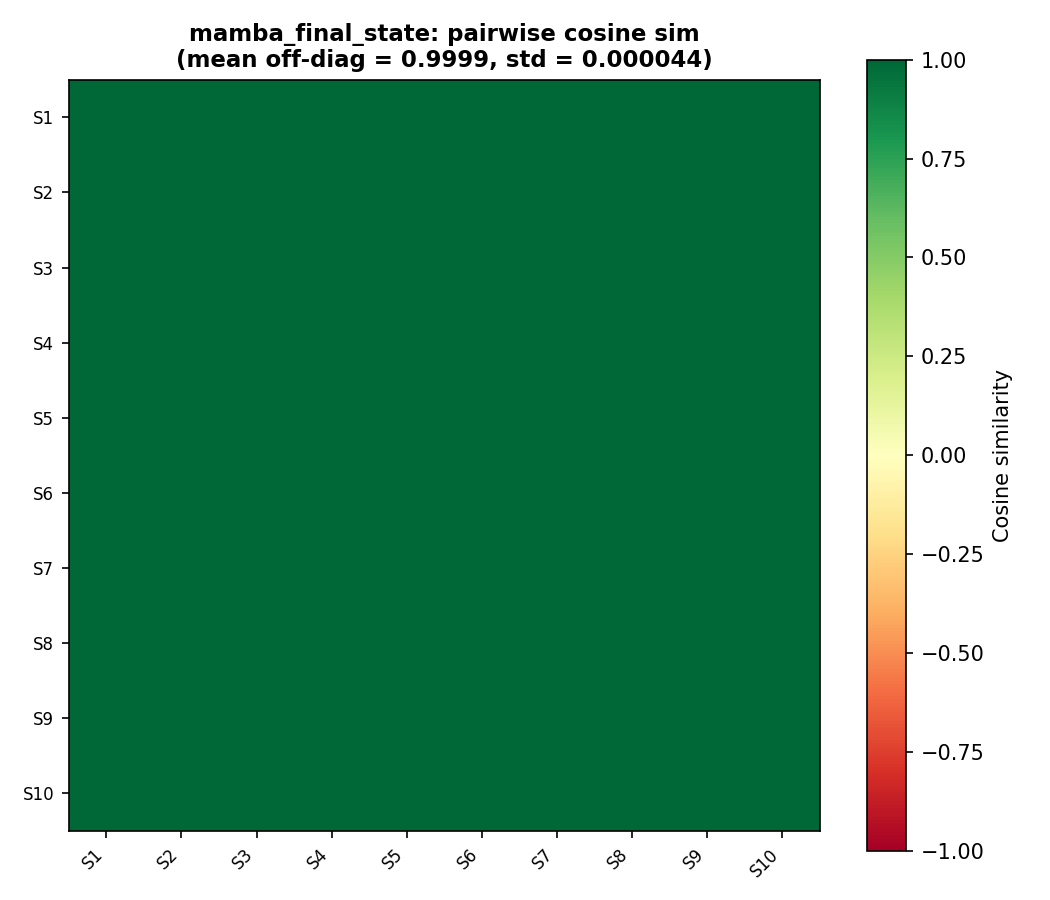}%
\hfill%
\includegraphics[width=0.49\columnwidth]{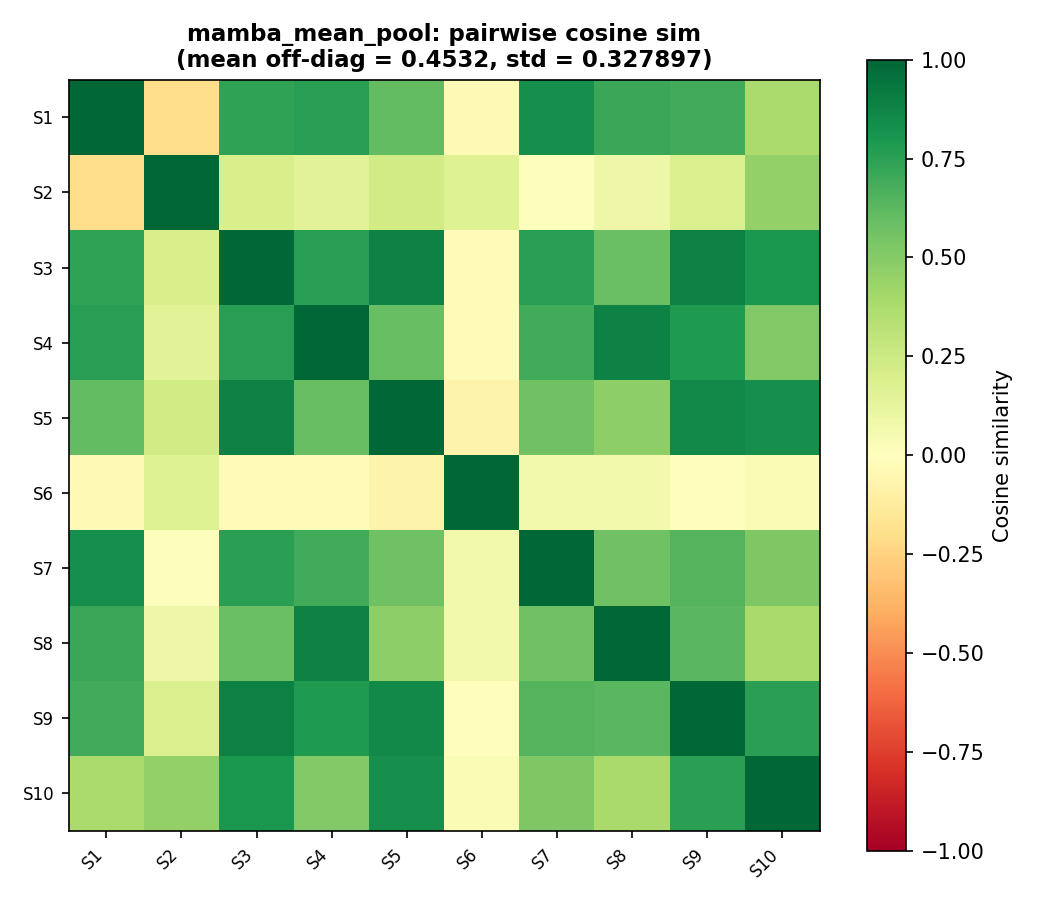}
\caption{%
\textbf{Pairwise cosine similarity heatmaps for ten semantically unrelated
sentences} (cats, quantum mechanics, stock markets, pizza, etc.).
\textit{Left:} \texttt{mamba\_final\_state}. Every off-diagonal cell is
the same dark green (mean $= 0.9999$, std $= 0.000044$, diagonal
excluded): all state vectors point in essentially the same direction
regardless of content.
\textit{Right:} \texttt{mamba\_mean\_pool} from the same backbone. The
matrix shows clear variation (mean $= 0.453$, std $= 0.328$).
Anisotropy is a property of the raw SSM state, not of the Mamba
backbone as a whole.%
}
\label{fig:heatmaps}
\end{figure}

Figure~\ref{fig:heatmaps} visualizes the anisotropy finding directly.
Ten sentences covering completely unrelated topics produce Final State
vectors that are indistinguishable in direction.
Inspecting the raw values makes this concrete: the first eight dimensions
for ``The cat sat on the mat'' and ``Quantum entanglement defies classical
intuition'' read as $[-0.0024,\ 0.0045,\ -0.0032,\ \ldots]$ and
$[-0.0024,\ 0.0045,\ -0.0033,\ \ldots]$ respectively, differing only
in the third or fourth decimal place across all dimensions.

The same backbone's mean-pooled token outputs show clear variation across
the same sentences, confirming that the pathology is specific to the raw
accumulated state and not an artifact of the model or tokenizer.

We hypothesize that this reflects the SSM weight structure.
The $\mathbf{B}$ projection matrix and the slow-decay $\mathbf{A}$ matrix
together bias all state accumulations toward a dominant direction in
$\mathbb{R}^D$.
Every sentence drives the state in approximately the same direction because
the projection operator is shared across all inputs.
Direct verification through spectral analysis of $\mathbf{B}$ or
layerwise ablation is left to future work.

BERT models are also anisotropic ($\bar{\rho}_\text{cos} \approx 0.99$).
The gap between $0.99$ and $0.9999$ is not cosmetic: angular deviation
is proportional to $\sqrt{1 - \cos\theta}$, which is roughly $10\times$
smaller at $0.9999$, severely degrading any metric that relies on angles.

\subsection{Why Orthogonal Injection Fails at Inference Time}

We evaluated Equation~\ref{eq:ortho} at fixed $\eta = 0.5$ with no
training on STS-B.
The result: Spearman $0.316 \to 0.259$, with mean cosine similarity
unchanged at $0.9999$.

The failure is expected in retrospect.
Anisotropy is a property of the weight matrices $\mathbf{A}_\text{log}$,
$\mathbf{B}$, and $\mathbf{C}$, not of the per-step write dynamics.
A fixed $\eta$ without gradient signal cannot overcome the directional
bias encoded in these weights.

Additionally, at the start of each sequence, $\mathbf{h}_0 = \mathbf{0}$,
so the projection term in Equation~\ref{eq:ortho} vanishes identically
for the first token.
Orthogonalization only activates once the state is already biased toward
its dominant direction.

Reducing anisotropy in Mamba likely requires training-time intervention:
contrastive fine-tuning to directly penalize anisotropy in the
representation space~\citep{gao2021simcse}, or training from scratch with
orthogonal injection active so that the weight matrices learn to exploit
the expanded representational geometry.
Post-hoc calibration methods developed for
transformers~\citep{mu2018allbutthetop} may also be applicable and are a
natural direction for follow-up.

\section{Related Work}

\paragraph{Mamba for NLP.}
Mamba~\citep{gu2023mamba} and Mamba-2~\citep{dao2024mamba2} have been
studied for language modeling and as components of hybrid
architectures~\citep{lieber2024jamba}.
Our work systematically probes the geometric quality of pretrained
Mamba's internal representations for sentence-level NLP under a
frozen-feature protocol, a dimension that has not been previously
characterized.

\paragraph{Sentence Embeddings.}
Extracting sentence representations from pretrained models is well-studied
for transformers~\citep{devlin2019bert,reimers2019sentencebert,gao2021simcse}.
SimCSE~\citep{gao2021simcse} addresses anisotropy in BERT through
contrastive fine-tuning, the kind of training-time intervention our
negative results suggest is also necessary for Mamba.
Our work extends this line of investigation to SSMs.

\paragraph{Probing.}
Frozen-feature probing as a framework for understanding what
representations encode was formalized in~\citet{alain2016understanding}
and applied extensively to
transformers~\citep{tenney2019bert,hewitt2019structural}.
Our three-seed multi-benchmark protocol provides statistical grounding,
and the CoLA collapse finding is a qualitatively strong failure mode tied
specifically to raw SSM state extraction.

\paragraph{Anisotropy.}
Anisotropy in word and sentence embeddings was characterized
in~\citet{ethayarajh2019contextual} and linked to training dynamics
in~\citet{gao2019representation}.
Post-hoc calibration methods~\citep{mu2018allbutthetop,li2020selfdot} can
reduce anisotropy in transformer representations; we leave their
application to Mamba states to future work.

\section{Conclusion}

We began with a mechanistically motivated hypothesis: that SSM recurrent
states at patch boundaries constitute useful, zero-cost sentence
representations.
Five benchmarks and a proposed recurrence modification later, we can say
more precisely why that hypothesis does not hold for this pretrained
Mamba-130M setting.

Our results suggest that the primary bottleneck is not extraction
strategy alone, but representational geometry.
Pretrained Mamba raw SSM states are severely anisotropic (mean cosine
similarity $0.9999$, std $0.000044$, measured over 1{,}000 sampled pairs
with the diagonal excluded).
This value does not appear to be a numerical artifact; our results
suggest it is tied to the geometry of the pretrained recurrence,
specifically the contraction properties of the A matrix and the shared
B projection operator accumulating writes in a dominant direction
across all inputs.
And without the learned output projection, a probe trained for ten full
epochs on the resulting constant-direction inputs cannot do better than
predict the majority class, which is what MCC $= 0.000$ means in
practice.

These findings are, we believe, useful to the community.
They establish quantified failure modes under frozen-feature probing and
point to the pretraining stage as the right place for intervention.
They also clarify what the output projection and patch boundaries
\emph{do} accomplish: they preserve information that is not linearly
accessible from the raw accumulated state, a structurally meaningful
contribution even when it does not produce a consistent downstream win.

\section*{Limitations}

All experiments use a single backbone (Mamba-130M), so the degree to
which these findings generalize across scales or pretraining recipes is
unknown.
SST-2 and IMDb results are single-seed pilots; CoLA and MRPC results are
more reliable.
The effect of patch size $P$ on the collapse phenomenon is an open
question.
The full method comparison on QQP was precluded by per-epoch compute
costs of approximately 80 minutes under the sequential Mamba fallback
required without CUDA kernel acceleration; partial results for
Patched-Mamba are reported in Section~\ref{sec:results}.
The geometric account of anisotropy offered here is a hypothesis; direct
verification through spectral analysis of $\mathbf{B}$ or layerwise
ablation is left to future work.

\section*{Acknowledgments}

Experiments were conducted on the University of Washington Hyak
high-performance computing cluster.
The authors thank the Hyak support team for infrastructure assistance.

\bibliographystyle{plainnat}
\bibliography{refs}

\end{document}